%% file: statement.tex
\documentclass[conference]{IEEEtran}

\input{custom_packages}
\input{custom_definitions}

\input{acronyms}

\usepackage{balance}
\begin{document}

\title{Safe and Efficient Estimation for Robotics \\Through the Optimal Use of Resources}


\author{\authorblockN{Frederike Dümbgen}
\authorblockA{Robotics Institute, University of Toronto, 
E-mail: frederike.dumbgen@utoronto.ca}} 

%

\maketitle

\IEEEpeerreviewmaketitle

\section{Introduction}

\begin{figure}[tb]
  \centering
  \includegraphics[width=\linewidth]{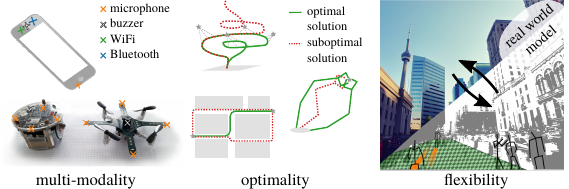}
  \caption{Three research directions to create more resilient robots by optimally using resources: using all available sensors (\textit{multi-modality}), using optimal solvers (\textit{optimality}), and learning models that can be adapted and solved efficiently (\textit{flexibility}).}
  \label{fig:models}
\end{figure}

In order to operate in and interact with the physical world, robots need to have estimates of the current and future state of the environment. We thus equip robots with sensors and build models and algorithms that, given some measurements, produce estimates of the current or future states. Environments can be unpredictable and sensors are not perfect. Therefore, it is important to both use all information available, and to do so optimally: making sure that we get the best possible answer from the amount of information we have. However, in prevalent research, uncommon sensors, such as sound or radio-frequency (RF) signals, are commonly ignored for state estimation~\cite{cadena_past_2016}; and the most popular solvers employed to produce state estimates are only of ``local'' nature, meaning they may produce suboptimal estimates for the typically non-convex estimation problems~\cite{rosen_se-sync_2019,yang_certifiably_2023}. My research aims to use resources more optimally, by building on the following three pillars, shown also in Figure~\ref{fig:models}.

\textbf{Multi-modality:} Many robotic platforms are equipped with sensors such as microphones and RF receivers, but their signals are not commonly used for state estimation.Using these sensing modalities as a complement can improve robustness against failure or inadequacy of more commonly used sensors such as cameras. I thus study and deploy uncommon sensors for state estimation, addressing their unique challenges such as finding accurate models and dealing with low signal-to-noise (SNR) ratios. 

\textbf{Optimality:} My second research focus is on designing solvers which exploit the measurements and designed models in an optimal and efficient way. In particular, I seek globally optimal solvers for non-convex, underdetermined, and low-SNR problems, by either adding optimality certificates (similar to ``quality badges'') to estimates, or designing solvers that are provably optimal. 

\textbf{Flexibility:} All models are wrong, so automating and facilitating the process of learning and improving models is of paramount importance. In this third direction of research, I investigate novel, transparent approaches to automatic problem formulation and model learning, taking into account the requirements of downstream optimization methods to learn not only accurate but also efficiently solvable models.

By making advances in these three research directions, we not only create more resource-efficient and resilient robots, but also push fundamental modeling and solving capabilities with reach beyond robotics.

\section{Past research}

\subsection{Multi-modality}\label{sec:novelty}

Arguably one of the most often deployed sensor modalities in robotics are cameras and lidar sensors~\cite{cadena_past_2016}. Their underlying physics are well understood and plenty appropriate models and algorithms have been developed~\cite{hartley_multiple_2004,zhao_method_2021}. However, not in all situations are such rich sensors the ideal choice; for example, when visibility is bad or when cost or weight are a concern~\cite{basiri_audio-based_2014}. 

For the application of indoor localization for museums and show rooms, for example, I have developed a system to localize a moving mobile phone~\cite{ipin} based on mostly Bluetooth and WiFi measurements, because cameras in this setting may pose privacy and usability issues. Because indoor RF signals exhibit low SNR, I created an algorithm that can auto-calibrate and output likelihood maps instead of single estimates. We also showed how using visual localization for auto-calibration, whenever available, greatly improves performance.

In a second application, I have investigated using sound for localization~\cite{dumbgen_blind_2023}, using light-weight speakers and microphones also commonly available on small robots~\cite{epuck}, as opposed to prior work which relies on considerably higher form factors. Using a small buzzer to emit frequency chirps and four microphones, and a simple image-source model for sound, the interference between direct-path and reflected-path waves can be used to accurately detect and localize a close-by wall. Given the high noise levels, I designed an algorithm that can efficiently fuse measurements over time and tested it on both the E-puck robot~\cite{epuck} and the Crazyflie drone with a custom-built audio deck~\cite{crazyflie}.

Both research projects have shown that these widely available and affordable sensing modalities, currently seldomnly utilized for state estimation, can complement more common modalities such as vision or lidar to improve estimation accuracy and resilience.

\subsection{Optimality}\label{sec:optimality}

In the former method, a state-of-the-art non-linear solver~\cite{kaess_isam2_2012} was deployed to solve for the locations of walls and the drone's trajectory. However, the optimization problem is non-convex and if initialized poorly, the solver may converge to a poor local minimum. In recent years, important progress has been made towards solvers that can deal with such non-convex problems. Using semidefinite relaxations of the original (polynomial) estimation problems, we can 
either solve the problem as a semidefinite program, or certify the solution of a standard local solver using Lagrangian duality~\cite{boyd_convex_2004}. This process is depicted in Figure~\ref{fig2}.

While prior work on using these principles in robotics was mostly limited to computer-vision applications~\cite{briales_certifiably_2018,yang_certifiably_2023} and pose-graph optimization~\cite{carlone_planar_2016,rosen_se-sync_2019}, me and my colleagues have extended them to certify general matrix-weighted localization and~\ac{SLAM}~\cite{holmes_semidefinite_2023}, for reprojection-error-based stereo camera localization and robust estimation~\cite{dumbgen_toward_2023}, range-only localization~\cite{dumbgen_safe_2023} and pose estimation~\cite{goudar_optimal_2024}, and finally, trajectory estimation using the Cayley map~\cite{barfoot_certifiably_2023}. We have thus extended the `catalogue' of problems that can be solved to global optimality, putting an emphasis on exploiting known sparsity patterns, incorporating motion priors, and evaluating on real-world data, thus ensuring the applicability to robotics problems.

\subsection{Flexibility}\label{sec:flexibility}

Appropriate solvers and problem formulations are interdependent aspects, and treating them in isolation may be suboptimal. In this third research direction, I investigate how to automatically find and update models and problem formulations which can be solved efficiently and optimally. 

Our first contribution in this direction consists of a method to semi-automatically determine a problem formulation which results in a tight semidefinite relaxation, a prerequisite for the methods described in the previous section. In many prior works, it was found that so-called redundant constraints are necessary in order to achieve this, and usually a tedious manual process is required to find the correct ones~\cite{ruiz_using_2011,yang_teaser_2020,yang_certifiably_2023,barfoot_certifiably_2023,holmes_semidefinite_2023}. We show that this process can in fact be automated by generating a sufficient number of feasible samples using the problem model, and determining the null-space of a corresponding data matrix~\cite{dumbgen_toward_2023}. This method allows for the ``tightening'' of a given formulation, but also for the incremental creation of a good formulation. We have applied this method to existing and novel problems, showing that it effortlessly creates tight relaxations with often fewer redundant constraints than previously thought, which considerably improves solver times.

\begin{figure}[tb]
  \centering
  \includegraphics[width=\linewidth]{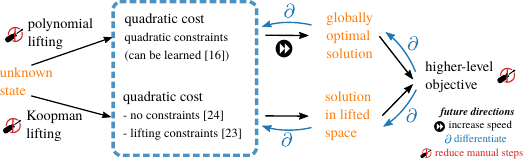}
  \caption{Overview of the used pipelines for modeling and solving for higher optimality and flexibility. The three main directions for future research are also depicted.}
  \label{fig2}
\end{figure}

In a parallel collaboration, we have leveraged Koopman operator theory~\cite{koopman_hamiltonian_1931} to lift a state into higher dimensions where its dynamics and measurement models become linear, depicted in Figure~\ref{fig2}. Exploiting linearity, globally optimal and efficient solvers can be derived in the lifted space~\cite{guo_data-driven_2023,guo_koopman_2022}. We relax the oftentimes used assumption that the state is measured directly~\cite{abraham_active_2019,mauroy_koopman_2020} and observe the state through non-linear measurements instead, which we also lift. We show that we can learn more accurate models than when using physics-based approaches, and correctly identify an unknown measurement bias for~\ac{UWB} measurements, for example. 

\balance 
\section{Ongoing and future research}

In ongoing research, I am pushing the \textbf{speed and scalability of globally optimal solvers}, which currently suffer from the cubic cost of solving a large semidefinite program, which means that most proposed solutions from Section~\ref{sec:optimality} are restricted to moderate-size problems. We use a feature ignored in these works, which is that the underlying problem structure is often chordally sparse, meaning that the problem can be decomposed into smaller, interconnected problems. In preliminary studies, I have seen that for problems without loop closures, we can reach linear complexity by decomposition and parallelizing. In the future, I hope to expand this method to non-chordal problems, using for example factor-width decompositions~\cite{zheng_chordal_2021} or loopy belief propagation~\cite{ortiz_visual_2021}. 

Secondly, me and my colleagues are improving the flexibility of models by using differentiable programming. Building on recent work in the machine learning and control community~\cite{agrawal_differentiable_2019,amos_differentiable_2019}, we can in fact \textbf{embed our optimal solvers in an end-to-end learned framework}, thus allowing for the problem parameters to be tuned to optimize a higher-level task, as depicted in Figure~\ref{fig2}. First conducted studies show that this approach is effective for tuning individual model parameters; the adoption of such an approach to more holistic tuning and model learning are essential follow-up problems that I aim to solve to further increase the flexibility and thus accuracy of the models. The same methods could also be used to incrementally update learned models in the Koopman framework in~\cite{guo_data-driven_2023}, making online adaptation of the models more efficient. 

My longer-term research goal is to enable the \textbf{effortless creation of data-driven but transparent models}. Essential to this endeavour is the removal of as many manual steps in the modeling process as possible, for which our prior work~\cite{dumbgen_toward_2023,guo_data-driven_2023} can be used as a stepping stone. My vision is to create a suite of tools rooted in rigorous mathematics, which researchers, notably not only roboticists, can use to automatically identify models from experimental data of different sensor modalities, and to efficiently and accurately solve for, and certify, state estimates. Both the developed methods, but also advances in theory that are required to make this vision a reality, have the potential to foster progress in all areas of robotics and more generally in safe autonomy.

\footnotesize{
\bibliographystyle{IEEEtran}
\bibliography{RSS}
}

\end{document}

%% file: custom_packages.tex
\usepackage{graphics} 
\usepackage{epsfig} 
\usepackage{times} 
\usepackage{amsmath} 
\usepackage{amssymb}  

\usepackage[textsize=footnotesize]{todonotes}

\usepackage{graphicx}
\usepackage{footnote}

\usepackage[ruled]{algorithm2e}

\usepackage{amsthm}
\usepackage{amsfonts}
\usepackage{mathtools}

\usepackage{tabularx,colortbl}
\usepackage{booktabs}
\usepackage[outdir=export/]{epstopdf}
\usepackage{multirow}
\usepackage{wrapfig}

\usepackage{algorithmic}
\usepackage{array}
\newcounter{rowcntr}[table]
\usepackage[font={small,it},belowskip=0pt]{caption}
\usepackage{placeins} 

\usepackage[percent]{overpic}
\usepackage{hyphenat}

\usepackage{tabu}
\usepackage{xcolor}
\usepackage{bm}
\usepackage{textcomp}

\usepackage[nolist,nohyperlinks]{acronym}
\usepackage{mathtools}

\usepackage{adjustbox}
\usepackage{float}
\usepackage{xspace}
\usepackage{empheq} 

\usepackage{siunitx}

%% file: custom_definitions.tex
\graphicspath{{drawings/}{export/}}

\DeclarePairedDelimiter\abs{\lvert}{\rvert}
\DeclarePairedDelimiter\norm{\lVert}{\rVert}
\makeatletter
\let\oldabs\abs
\let\oldnorm\norm
\def\abs{\@ifstar{\oldabs}{\oldabs*}}
\def\norm{\@ifstar{\oldnorm}{\oldnorm*}}
\makeatother

\def\-{\raisebox{0pt}{-}}

\newtheoremstyle{example}%
{}
{}
{\itshape}
{}
{\bfseries}
{\normalfont{.}}
{ }
{}
\newtheoremstyle{definition}%
{}
{}
{\itshape}
{}
{\bfseries}
{\normalfont{.}}
{ }
{}
\theoremstyle{definition}

\theoremstyle{example}

\let\oldunderbrace\underbrace
\renewcommand{\underbrace}[2]{\let\scriptstyle\textstyle \oldunderbrace{#1}_{#2}}
\definecolor{tab10blue}{rgb}{0.12156863, 0.46666667, 0.70588235}

\DeclareFontFamily{U}{mathx}{\hyphenchar\font45}
\DeclareFontShape{U}{mathx}{m}{n}{ <-> mathx10 }{}
\DeclareSymbolFont{mathx}{U}{mathx}{m}{n}
\DeclareFontSubstitution{U}{mathx}{m}{n}
\DeclareMathAccent{\widebar}{\mathalpha}{mathx}{"73}
\newcolumntype{C}[1]{>{\centering\let\newline\\\arraybackslash\hspace{0pt}}m{#1}}
\newcolumntype{D}[1]{>{\centering\let\newline\\\arraybackslash\hspace{0pt}\columncolor{gray!20}}m{#1}}
\newcolumntype{R}[1]{>{\raggedleft\let\newline\\\arraybackslash\hspace{0pt}}m{#1}}
\newcolumntype{L}[1]{>{\raggedright\let\newline\\\arraybackslash\hspace{0pt}}m{#1}}
\newcolumntype{M}[1]{>{\raggedright\let\newline\\\arraybackslash\hspace{0pt}}m{#1}}
\newcolumntype{N}{>{\refstepcounter{rowcntr}\therowcntr}c}

\usepackage{tabstackengine}
\stackMath
\setstackgap{L}{18pt}
\setstacktabbedgap{0pt}
\def\lrgap{\kern6pt}

  \renewenvironment{thebibliography}[1]{%
    \begin{oldthebibliography}{#1}%
      \setlength{\parskip}{0ex}%
      \setlength{\itemsep}{0ex}%
  }%
  {%
    \end{oldthebibliography}%
  }


%% file: acronyms.tex
\acrodef{QCQP}[QCQP]{quadratically constrained quadratic program}
\acrodef{SDP}[SDP]{semidefinite program}
\acrodef{EVD}[EVD]{eigenvalue decomposition}
\acrodef{SVD}[SVD]{singular value decomposition}
\acrodef{MAP}[MAP]{maximum a posteriori}
\acrodef{GP}[GP]{Gaussian process}
\acrodef{LTV}[LTV]{linear, time-varying}
\acrodef{SLAM}[SLAM]{simultaneous localization and mapping}
\acrodef{IMU}[IMU]{inertial measurement unit}
\acrodef{psd}[PSD]{positive-semidefinite}
\acrodef{GN}[GN]{Gauss-Newton}
\acrodef{UWB}[UWB]{ultra-wideband}
\acrodef{rmse}[RMSE]{root-mean-squared error}
\acrodef{mae}[MAE]{mean absolute error}
\acrodef{LM}[LM]{Levenberg-Marquardt}
\acrodef{GPS}[GPS]{global positioning system}
\acrodef{AUV}[AUV]{autonomous underwater vehicle}
\acrodef{SOS}[SOS]{sums-of-squares}
\acrodef{RO}[RO]{range-only}
\acrodef{KKT}[KKT]{Karush-Kuhn-Tucker}
\acrodef{TLS}[TLS]{truncated least-sqares}
\acrodef{SVR}[SVR]{singular-value ratio}
\acrodef{POP}[POP]{polynomial optimization problem}
\acrodef{RDG}[RDG]{relative duality gap}
\acrodef{MPC}[MPC]{model predictive control}
\acrodef{SQP}[SQP]{sequential quadratic programming}
\acrodef{QP}[QP]{quadratic program}
\acrodef{AI}[AI]{artificial intelligence}
\acrodef{IK}[IK]{inverse kinematics}
\acrodef{DOF}[DoF]{degree of freedom}
\acrodef{NERF}[NERF]{neural radiance fields}
\acrodef{MPC}[MPC]{model-predictive control}
\acrodef{RL}[RL]{reinforcement learning}
\acrodef{HJ}[HJ]{Hamilton-Jacobi}
\acrodef{CBF}[CBF]{control barrier function}
\acrodef{CLF}[CLF]{control Lyapunov function}
\acrodef{UAV}[UAV]{unmanned aerial vehicle}
\acrodef{RF}[RF]{radio frequency}
\acrodef{LIDAR}[lidar]{light detection and ranging}
\acrodef{RTT}[RTT]{round-trip time}
\acrodef{RSS}[RSS]{received signal strength}

%% file: statement.bbl
\begin{thebibliography}{10}
\providecommand{\url}[1]{}
\csname url@rmstyle\endcsname
\providecommand{\newblock}{\relax}
\providecommand{\bibinfo}[2]{#2}
\providecommand\BIBentrySTDinterwordspacing{\spaceskip=0pt\relax}
\providecommand\BIBentryALTinterwordstretchfactor{4}
\providecommand\BIBentryALTinterwordspacing{\spaceskip=\fontdimen2\font plus
\BIBentryALTinterwordstretchfactor\fontdimen3\font minus
  \fontdimen4\font\relax}
\providecommand\BIBforeignlanguage[2]{{%
\expandafter\ifx\csname l@#1\endcsname\relax
\typeout{** WARNING: IEEEtran.bst: No hyphenation pattern has been}%
\typeout{** loaded for the language `#1'. Using the pattern for}%
\typeout{** the default language instead.}%
\else
\language=\csname l@#1\endcsname
\fi
#2}}

\bibitem{cadena_past_2016}
C.~Cadena, L.~Carlone, H.~Carrillo, Y.~Latif, D.~Scaramuzza, J.~Neira, I.~Reid,
  and J.~J. Leonard, ``{Past, Present, and Future of Simultaneous Localization
  and Mapping: {{Toward}} the Robust-Perception Age},'' \emph{IEEE Transactions
  on Robotics}, vol.~32, no.~6, pp. 1309--1332, 2016.

\bibitem{rosen_se-sync_2019}
D.~M. Rosen, L.~Carlone, A.~S. Bandeira, and J.~J. Leonard, ``{SE-sync: A
  Certifiably Correct Algorithm for Synchronization over the Special Euclidean
  Group},'' \emph{International Journal of Robotics Research}, vol.~38, no.
  2-3, pp. 95--125, 2019.

\bibitem{yang_certifiably_2023}
H.~Yang and L.~Carlone, ``Certifiably {{Optimal Outlier-Robust Geometric
  Perception}}: {{Semidefinite Relaxations}} and {{Scalable Global
  Optimization}},'' \emph{IEEE Transactions on Pattern Analysis and Machine
  Intelligence}, vol.~45, no.~3, pp. 2816--2834, 2023.

\bibitem{hartley_multiple_2004}
R.~Hartley and A.~Zisserman, \emph{Multiple {{View Geometry}} in {{Computer
  Vision}}}, 2nd~ed.\hskip 1em plus 0.5em minus 0.4em\relax Cambridge
  University Press, 2004.

\bibitem{zhao_method_2021}
J.~Zhao, Y.~Li, B.~Zhu, W.~Deng, and B.~Sun, ``Method and {{Applications}} of
  {{Lidar Modeling}} for {{Virtual Testing}} of {{Intelligent Vehicles}},''
  \emph{IEEE Transactions on Intelligent Transportation Systems}, vol.~22,
  no.~5, pp. 2990--3000, 2021.

\bibitem{basiri_audio-based_2014}
M.~Basiri, F.~Schill, D.~Floreano, and P.~U. Lima, ``Audio-based localization
  for swarms of micro air vehicles,'' in \emph{{{IEEE International
  Conference}} on {{Robotics}} and {{Automation}} ({{ICRA}})}, 2014, pp.
  4729--4734.

\bibitem{ipin}
F.~D{\"u}mbgen, C.~Oeschger, M.~Kolundzija, A.~Scholefield, E.~Girardin,
  J.~Leuenberger, and S.~Ayer, ``Multi-modal probabilistic indoor localization
  on a smartphone,'' in \emph{International {{Conference}} on {{Indoor
  Positioning}} and {{Indoor Navigation}} ({{IPIN}})}, 2019.

\bibitem{dumbgen_blind_2023}
F.~D{\"u}mbgen, A.~Hoffet, M.~Kolund{\v z}ija, A.~Scholefield, and M.~Vetterli,
  ``Blind as a {{Bat}}: {{Audible Echolocation}} on {{Small Robots}},''
  \emph{IEEE Robotics and Automation Letters}, vol.~8, no.~3, pp. 1271--1278,
  2023.

\bibitem{epuck}
F.~Mondada, M.~Bonani, X.~Raemy, J.~Pugh, C.~Cianci, A.~Klaptocz, J.-C.
  Zufferey, D.~Floreano, and A.~Martinoli, ``The e-puck, a robot designed for
  education in engineering,'' in \emph{Proceedings of the 9th Conference on
  Autonomous Robot Systems and Competitions}, 2009, pp. 59--65.

\bibitem{crazyflie}
W.~Giernacki, M.~Skwierczy{\'n}ski, W.~Witwicki, P.~Wro{\'n}ski, and
  P.~Kozierski, ``{Crazyflie 2.0 Quadrotor as a Platform for Research and
  Education in Robotics and Control Engineering},'' in \emph{22nd
  {{International Conference}} on {{Methods}} and {{Models}} in {{Automation}}
  and {{Robotics}}}, 2017, pp. 37--42.

\bibitem{kaess_isam2_2012}
M.~Kaess, H.~Johannsson, R.~Roberts, V.~Ila, J.~J. Leonard, and F.~Dellaert,
  ``{iSAM2: Incremental Smoothing and Mapping Using the Bayes Tree},''
  \emph{International Journal of Robotics Research}, vol.~31, no.~2, pp.
  216--235, 2012.

\bibitem{boyd_convex_2004}
S.~Boyd and L.~Vandenberghe, \emph{Convex {{Optimization}}}.\hskip 1em plus
  0.5em minus 0.4em\relax Cambridge University Press, 2004.

\bibitem{briales_certifiably_2018}
J.~Briales, L.~Kneip, and J.~{Gonzalez-Jimenez}, ``A {{Certifiably Globally
  Optimal Solution}} to the {{Non-minimal Relative Pose Problem}},'' in
  \emph{{{IEEE}}/{{CVF Conference}} on {{Computer Vision}} and {{Pattern
  Recognition}} ({{CVPR}})}, 2018, pp. 145--154.

\bibitem{carlone_planar_2016}
L.~Carlone, G.~C. Calafiore, C.~Tommolillo, and F.~Dellaert, ``Planar {{Pose
  Graph Optimization}}: {{Duality}}, {{Optimal Solutions}}, and
  {{Verification}},'' \emph{IEEE Transactions on Robotics}, vol.~32, no.~3, pp.
  545--565, 2016.

\bibitem{holmes_semidefinite_2023}
C.~Holmes, F.~D{\"u}mbgen, and T.~D. Barfoot, ``On {{Semidefinite Relaxations}}
  for {{Matrix-Weighted State-Estimation Problems}} in {{Robotics}},''
  \emph{arXiv:2308.07275 [cs, math]}, 2023.

\bibitem{dumbgen_toward_2023}
F.~D{\"u}mbgen, C.~Holmes, B.~Agro, and T.~D. Barfoot, ``Toward {{Globally
  Optimal State Estimation Using Automatically Tightened Semidefinite
  Relaxations}},'' \emph{arXiv:2308.05783 [cs]}, 2023.

\bibitem{dumbgen_safe_2023}
F.~D{\"u}mbgen, C.~Holmes, and T.~D. Barfoot, ``Safe and {{Smooth}}:
  {{Certified Continuous-Time Range-Only Localization}},'' \emph{IEEE Robotics
  and Automation Letters}, vol.~8, no.~2, pp. 1117--1124, 2023.

\bibitem{goudar_optimal_2024}
A.~Goudar, F.~D{\"u}mbgen, T.~D. Barfoot, and A.~P. Schoellig, ``Optimal
  {{Initialization Strategies}} for {{Range-Only Trajectory Estimation}},''
  \emph{IEEE Robotics and Automation Letters}, vol.~9, no.~3, pp. 2160--2167,
  2024.

\bibitem{barfoot_certifiably_2023}
T.~D. Barfoot, C.~Holmes, and F.~D{\"u}mbgen, ``Certifiably {{Optimal
  Rotation}} and {{Pose Estimation Based}} on the {{Cayley Map}},''
  \emph{arXiv:2308.12418 [cs]}, 2023.

\bibitem{ruiz_using_2011}
J.~P. Ruiz and I.~E. Grossmann, ``{Using Redundancy to Strengthen the
  Relaxation for the Global Optimization of {{MINLP}} Problems},''
  \emph{Computers \& Chemical Engineering}, vol.~35, no.~12, pp. 2729--2740,
  2011.

\bibitem{yang_teaser_2020}
H.~Yang, J.~Shi, and L.~Carlone, ``{TEASER : Fast and Certifiable Point Cloud
  Registration},'' \emph{IEEE Transactions on Robotics}, vol.~32, no.~2, pp.
  314--333, 2020.

\bibitem{koopman_hamiltonian_1931}
B.~O. Koopman, ``Hamiltonian {{Systems}} and {{Transformation}} in {{Hilbert
  Space}},'' \emph{Proceedings of the National Academy of Sciences}, vol.~17,
  no.~5, pp. 315--318, 1931.

\bibitem{guo_data-driven_2023}
Z.~C. Guo, F.~D{\"u}mbgen, J.~R. Forbes, and T.~D. Barfoot, ``Data-{{Driven
  Batch Localization}} and {{SLAM Using Koopman Linearization}},''
  \emph{arXiv:2309.04375 [cs]}, 2023.

\bibitem{guo_koopman_2022}
Z.~C. Guo, V.~Korotkine, J.~R. Forbes, and T.~D. Barfoot, ``Koopman
  {{Linearization}} for {{Data-Driven Batch State Estimation}} of
  {{Control-Affine Systems}},'' \emph{IEEE Robotics and Automation Letters},
  2022.

\bibitem{abraham_active_2019}
I.~Abraham and T.~D. Murphey, ``Active {{Learning}} of {{Dynamics}} for
  {{Data-Driven Control Using Koopman Operators}},'' \emph{IEEE Transactions on
  Robotics}, vol.~35, no.~5, pp. 1071--1083, 2019.

\bibitem{mauroy_koopman_2020}
A.~Mauroy, I.~Mezi{\'c}, and Y.~Susuki, Eds., \emph{The {{Koopman Operator}} in
  {{Systems}} and {{Control}}: {{Concepts}}, {{Methodologies}}, and
  {{Applications}}}, ser. Lecture {{Notes}} in {{Control}} and {{Information
  Sciences}}.\hskip 1em plus 0.5em minus 0.4em\relax Springer International
  Publishing, 2020, vol. 484.

\bibitem{zheng_chordal_2021}
Y.~Zheng, G.~Fantuzzi, and A.~Papachristodoulou, ``{Chordal and Factor-Width
  Decompositions for Scalable Semidefinite and Polynomial Optimization},''
  \emph{Annual Reviews in Control}, vol.~52, pp. 243--279, 2021.

\bibitem{ortiz_visual_2021}
J.~Ortiz, T.~Evans, and A.~J. Davison, ``A visual introduction to {{Gaussian
  Belief Propagation}},'' \emph{arXiv:2107.02308 [cs]}, 2021.

\bibitem{agrawal_differentiable_2019}
A.~Agrawal, B.~Amos, S.~Barratt, and S.~Boyd, ``Differentiable {{Convex
  Optimization Layers}},'' in \emph{{{NeurIPS}}}, 2019.

\bibitem{amos_differentiable_2019}
B.~Amos, I.~Jimenez, J.~Sacks, B.~Boots, and J.~Z. Kolter, ``Differentiable
  {MPC} for end-to-end planning and control,'' \emph{{Advances in Neural
  Information Processing Systems}}, vol.~31, 2018.

\end{thebibliography}
